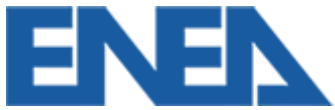

Italian National Agency for New Technologies, Energy and Sustainable Economic Development

# Determination of spatial configuration of an underwater swarm with minimum data

Ramiro dell'Erba

**Abstract** This paper is the extension of work presented at the IARP Conference "Bio inspired robotics" held in Frascati (Italy), 14 May 2014.

The subject is the localization problem of an underwater swarm of autonomous underwater robots (AUV), in the frame of the HARNESS project; by localization, we mean the relative swarm configuration, i.e., the geometrical shape of the group. The result is achieved by using the signals that the robots exchange. The swarm is organized by rules and conceived to perform tasks, ranging from environmental monitoring to terrorism-attack surveillance.

Two methods of determining the shape of the swarm, both based on trilateration calculation, are proposed. The first method focuses on the robot's speed. In this case, we use our knowledge of the speeds and distances between the machines, while the second method considers only distances and the orientation angles of the robots. Unlike a trilateration problem, we do not know the position of the beacons and this renders the problem a difficult one. Moreover, we have very few data. More than one step of motion is needed to resolve the multiple solutions found, owing to the symmetries of the system and optimization process of one or more objective functions leading to the final configuration. We subsequently checked our algorithm using a simulator taking into account random errors affecting the measurements

## 1. Introduction

This work was realized in the context of the HARNESS

project (Human telecontrolled Adaptive Robotic Network of SensorS) currently in progress in our laboratory [1], [2]. The HARNESS project aims at the realization of an underwater multi-body robotics system able to perform tasks in a fast and reliable way, and with a swarm-like behaviour. The availability of a suitable robotic swarm could be relevant in many operations: such as surveillance of sensitive sites, fast exploration of relatively interesting areas and detailed analysis of objects without removing them from their underwater sites (i.e., archaeological goods).

The aim of this work is the reconstruction of the swarm configuration using the signals they exchange. We have treated this task in [3] and [4], but now we have a new method. In the underwater world, a severe limitation of our communications technology is, perhaps, the main drawback: the physical medium only permits acoustic channels, since electromagnetic waves are rapidly dampened. The acoustic technology is heavily affected by a fast decay, a temporal delay, and so on, in the signal band-pass as soon as the distance increases, as well as in the case of limited ranges. This limitation can be overcome by a suitable and intelligent spatial distribution of transmission nodes (the swarm members themselves), allowing an enhanced throughput of data through logical and physical routing. In our system, an attempt to use a cheap optical communication system, together with the acoustic one, is under investigation.

Note that we are talking about the geometric arrangement of robots in space; therefore, getting an absolute localization for the group requires one member of the swarm to have an absolute localization: for example, when one member has a fixed position by GPS.

In this paper, we present the results of some calculations relative to possible methods that can be adopted in our system. This paper is an extension of a conference communication [4], the principal differences of which are described below. We present another algorithm based on the minimization of one or more objective functions; in the previous paper, we solved a system of equations that is less efficient. We present some approximation possibilities to obtain the configuration more quickly. Constraints placed on the variables to facilitate convergence of the calculations are introduced. We propose the possibility of working by integer numbers' approximation, taking into account the precise degree required; this ensures that we obtain an absolute minimum of the objective function that has the mean of the real configuration. Another software simulator was built. Random errors in the simulated measurements of the distances and orientation angles used for the calculation were introduced, and their influence evaluated. A large bibliography was added. We go on to describe the equipment that the robots must carry on board. Some possible operative scenarios are described. In conclusion, the objective of the two papers is the same: to obtain the configuration of the swarm. New results obtained include a new algorithm closer to real operating conditions, which work more efficiently.

In localizing the configuration of the swarm, we are dealing with something similar to a trilateration problem, but in our case, we do not know the position of the beacons; this makes the problem more difficult. Therefore, we try to obtain the configuration despite this circumstance and also that of a lack of data.

The calculation is based on trilateration between three or more robots in different steps of motion. The minimum data used in the calculation was the orientation of the robots and the distance between them.

The HARNESS project seeks to realize an underwater multi-AUV robotic system, arranged in a swarm organization where the classical flocking rules and the communication network protocol are merged in a novel, higher-level control. It is expected to improve the performance of classical AUV technology, exploiting the large occupied volume and the short distances between the vessels. The speed of volume monitoring and the transmission band-pass among the vessels and towards the surface should be some of the most important results, with respect to a single AUV system. The use of underwater autonomous vessels or rovers has been tested and proven to be useful in many cases, but generally expensive, mainly because it requires the support of an equipped ship. An AUV could be considered a cost-effective alternative to other available technologies, such as manned submersibles, remotely operated vehicles (ROVs) and towed instruments led by ships. However, many problems still need to be solved in order to make

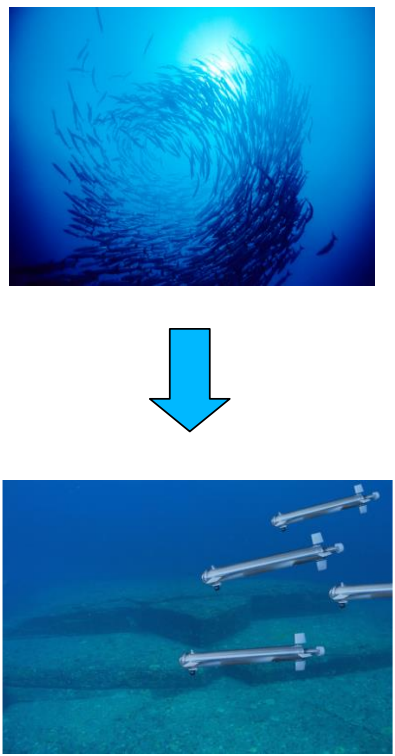

**Figure 1.** The basic idea of the HARNESS project.

AUVs competitive, especially for the issues relevant to power availability, information processing, navigation and control as explained in [5], [6], [7], [8], [9], [10].

A swarm is able to perform tasks in a quicker and more robust manner than is possible by a single machine [11], [12]. Moreover, a swarm has the advantage of the simplicity of interfacing with human users, which factor overcomes the problem of controlling a large number of individuals. In a swarm, the robots operate with a common objective and share the job workload. The problems caused by the absence of one member can be easily solved by redistributing the job among the remaining members, such as in natural systems like that of bees [13], [14], [15], [16], [17]. This feature is especially useful if we consider its potential application in the discovery and surveillance of submarine archaeological sites. A characteristic of the swarm is that the geometric distribution of this system's members is flexible and adaptable to the task and environment.

The most interesting areas to explore and protect are those in close proximity to coasts, and with depths ranging between 50 and 200 m. Professional and expensive divers can operate only to a maximum of 70-80 m in recovery operations, and simple and fast explorations cannot be performed by humans beyond 100 m.

The use of robotic technologies in ocean surveys, inspections, pipe and cable tracking has been well-established in the field of marine engineering for many years [10]. The performance of marine robotics technology has developed quickly in recent years, when many autonomous underwater vehicle systems moved from the prototype stage to scientific, commercial and military uses.

The goal of this paper is to present some of the novel concepts developed in the HARNESS project. We will discuss the advantages and drawbacks of the project's concepts and describe the prototype under construction. Finally, we will focus on the crucial problem of the localization of the swarm, offering some proposals for an efficient solution suitable in some scenarios.

## 2. The swarm concept

The use of a large number of very low-cost mini-AUVs [3], [9], could limit the use of the expensive surface ships to the deployment phase, taking advantage of the parallel exploration to shorten times and obtain many other advantages [6].

The concept of robot swarms has been a theme of study in the scientific community for several years. The realization of swarms of different numbers of cooperating robots has been successfully attempted, but is still a challenge in an underwater environment [18], [19], [20], [21], [22], [23], [24], [25]. Many of the difficulties of the unfriendly underwater environment are mainly focused on the problem of fast and reliable communication links. Swarm research has been inspired by biological behaviours, such as that of bees [15], [14], and has long taken advantage of concepts of social activities [26], labour division, task cooperation [27] and information sharing. A single-robot approach is affected by failures that may prevent the success of the whole task. However, a multi-robot approach can benefit by the parallelism of the operation, and by the redundancy of the use of multiple agents [28].

An advantage of the swarm, considered as a whole entity, lies in the possibility of parallelizing very heavy computation. As an example, an image analysis to recognize a landmark is a typical task that can be parallelized and distributed to the various machines, providing that an adequate communication network is available. In the same way, it offers the advantage of a simple means of interfacing with the human end-users, overcoming the problem of controlling a large number of individuals. On the other hand, the disadvantage is that a swarm needs one more layer in the control system: the layer that shares the task between the human and the individual machine [19].

In a swarm, the members operate with a common objective, sharing the job workload; the absence of one member can be easily taken care of by redistributing the job among the others. The geometrical distribution of the members of this system is flexible and adaptable to both the task's and the environment's characteristics; in particular, that of communication. In the underwater world, the physical medium makes the acoustic channels the most convenient ones, since electromagnetic waves are rapidly dampened [29], [30], [31]. The acoustic technology has limited performance; the band-pass increases as frequency increases, but its dampening limits the useful range [32]. The swarm technology allows us to avoid this drawback by means of a suitable and intelligent spatial distribution of transmission nodes [33], [34], with the swarm members themselves modifying the physical dispersion/geometry. This can be achieved by adapting and allowing the exploitation of ultra-high frequencies and an enhanced data transmission through logical and physical routing (See Fig. 2 and Fig. 3, where two examples of swarm configuration are shown for different tasks), [35], [36], [37], [21], [38], [39], [40], [41],

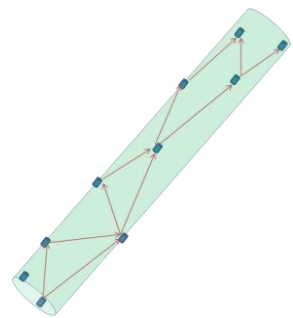

**Figure 2.** "Pipe" configuration, used when communication is the main objective of the geometrical shape, to transport data over long distances at the maximum allowed speed.

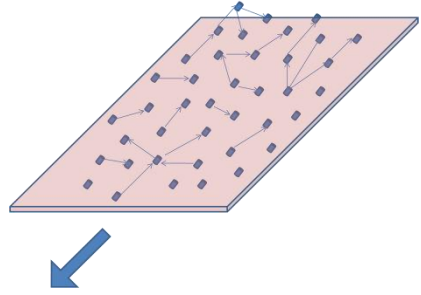

**Figure 3.** Planar distribution used to carry out fast and parallel monitor operations.

[33], [42]. In practise, it is a multi-hop network with varying geometry. Different geometry configurations of the nodes and the elements of the swarm itself, allow higher or lower data rate transmission. The use of a high frequency allows a higher band-pass, but it is well known that it is more dampened by the water. So far, an intelligent configuration of the node, as indicated in the figures, is desirable, varying the distances between them as a function of the task. One of the aims of the project is the study and implementation of different behaviours in the swarm, to generate a collective shaping as a response to environmental stimuli and to modify the communication parameters in order to maximize the system's performance.

The swarm control must balance three different objects: the various requests of the operator (e.g., modify the mission task), the swarm's needs and the management of its individual members (e.g., obstacle avoidance, loss of communication links, etc.).

This equilibrium can change depending on the assigned tasks, the survival risk associated with the operation of each robot and the risk associated with the loss of connection with the rest of the swarm. Approaches to these aims currently under study include neural network techniques, fuzzy logic and genetic algorithms [43]. The result is the selection of collective behaviours that must be compatible with all of the aforementioned conditions.

In the swarm, there is no central brain, mainly because of the excess needs of band-pass required by such a brain to communicate with all members of the swarm. Each individual must possess an intelligent local control system capable of managing its choices according to the choices of its neighbours, based on the available data. These data will differ depending on the position of the individual machine inside the swarm, the data propagation speed and, in special cases, by the role of some individuals (e.g., a scout going onto the surface for communication and localization by GPS).

Data coherence along the swarm, being affected by the position of the members and by the data propagation speed, is also a current research topic.

Therefore, the system has to adapt itself to the environmental characteristics, with special reference to the problems arising in communication within the swarm, and between the swarm and humans. These aspects are related to both the communication channel and to the geometrical distribution of the multi-robot system.

This can be obtained by a correct geometrical configuration of the swarm. In the case of swarm-human communication, an element of the swarm has to emerge in order to maintain the link with the other elements.

## 3.  The technological challenge

In this section, we offer some key concepts that we use to develop our swarm. A possible, fairly efficient way to achieve this result is to organize a number of cheap, small units able to change their relative cooperation modes, geometries and other functional parameters by means of relatively simple rules that every member of the swarm must follow; the Reynolds's boids [44] are a very famous example. Typical objectives of these rules are the maintenance of group coherence, the capability to follow a common target, the ability to avoid foreseen or unforeseen obstacles, and the ability to share and use data perceived by the entire group membership. All these group behaviours allow, in an implicit manner usually called "emergent behaviour", for the most appropriate means of interacting with the surrounding environment. Four topics are identified as key elements of the architecture: Communication, Control, Localization and Teleoperation. Each of those topics alone would keep many people busy for a long time, so we shall just offer a brief consideration of their application in the HARNESS project.

Communication distinguishes at least three different situations:
1) Fast data transmission rate over relatively long distances (High Speed Transmission);
2) The need to increase the swarm's internal data exchange (High Swarm Band-Pass. The geometry is presumably disordered);
3) The need to maintain a dispersed swarm configuration to accomplish wide-area monitoring tasks (Wide Area Surveillance). In this case, a Low Swarm Band-Pass is probably acceptable.

It is well known that from the swarm control level and the local individual control level, conflicting situations relevant to the optimization of global and local goals can originate. The choice of the right strategy to manage and synchronize this multilevel adaptive system is still under investigation and will be explored in greater detail during future developments of the project. It is sufficient, however, to state that it clearly involves the definition of rules in explicit (expert system) or implicit (neural networks) forms, and should be capable of on-the-job learning.

Three main control layers are a possible schematization of the control system: the Communication control, the

Swarm control and the Individual control. In addition, a fourth layer — the Arbiter — is deputed to solve conflicts that can arise among the previous levels. While communication and individual control are often studied, we underline the importance of the Swarm control and "Swarm rules", which are assigned but which sometimes have to be changed by the operator or as a result of the swarm's own learning.

The global control architecture will be built around three basic issues: the human supplied goals; the internal operational rules, ensuring stability and coordination of the members' behaviour through the communication network; and the emergencies and automatic answers of individuals (i.e., protection of themselves, obstacles avoidance).

The localization can be divided in three categories. Precise localization (i.e., in the arbour using a fixed buoy), Rough localization (i.e., during navigation), and Relative localization (i.e., geometric configuration of the swarm). We also used three scenarios: Structured environment, Free navigation and Close to objective. We shall go into the details of Relative localizations in the following paragraphs.

It is a difficult task to merge teleoperation with the swarm concept; therefore, we prefer the term 'TeleCooperation' to that of 'Teleoperation'. Orders from a human operator are passed on as needs with high priority, in accordance with the philosophy of the swarm. The swarm readapts on the basis of the new needs.

## 4. Our prototype

Figures 4 and 5 show the Venus prototype, realized in our laboratory. It has the following characteristics:

Maximum depth 100 m; Maximum speed 4 Km/hr; Weight approximately 20 Kg; Autonomy 3 hrs; Dimensions 1.20 m X 0.20 m in diameter.

Standard sensors include a stereoscopic camera, sonar, an accelerometer, a compass, a depth meter, hydrophones and side-scan sonar equipment.

We are dealing with a system made up of a swarm comprising 20 objects. Distances between robots are between 3 and 50 m. Therefore, the maximum possible distance between two robots is about 1000 m, as a very particular alignment case; the average value of the distances is approximately 10 m.

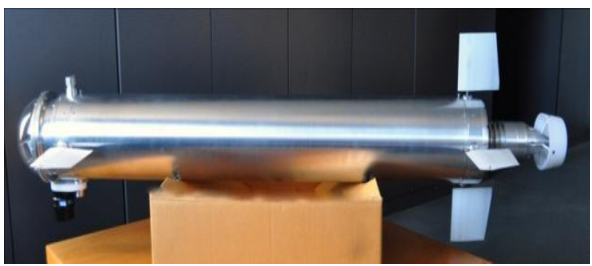

**Figure 4.** Low-cost Venus AUV.

It is our intention to use an optical, high-power transmission device together with the acoustic modem. It will be used for a number of different experimental approaches integrating the acoustic data channel and direct vision sensing. Optical methods are very powerful but their performance is affected by many very variable parameters such as salinity, turbidity, the presence of dissolved substances that change colour and the degree of transparency in different optical bands, and the amount of solar radiation, which has a strong effect on the signal to noise ratio. The current approach uses a mixed strategy based on the variable exploitation of the optical channel, depending on the environmental conditions. In favourable conditions, the transmission protocol will freely decide which channel to adopt, depending on the priority, distance-to-cover and dimension of the message itself. In less favourable conditions, the optical channel will be limited to the fundamental synchronization task, generating a light lamp that will optimize the message passing through the optical channel (and several other communication functions). In poor optical conditions, strong but very short time lamps will ensure references that allow for safe visual navigation. Therefore, communications can take place with greater or lower speed in the optical or acoustic domains, with different delays, attenuations, angular distributions of the radiated power, and so forth. We shall use one or the other depending on the environmental conditions: in fresh and homogeneous water, optical methods will be preferred.

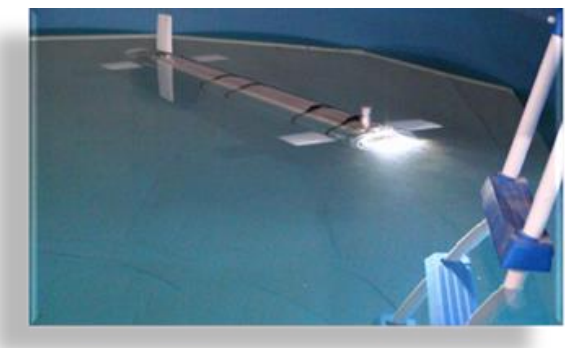

**Figure 5.** Robot prototype during a test in the pool.

## 5. The localization problem

Localization and (eventually) mapping are key to successful navigation in autonomous mobile platform technology, and are fundamental tasks performed in order to achieve high levels of autonomy in robot navigation, and robustness in vehicle positioning and values of the data [45], [46], [47], [48], [49], [50], [51], [52], [53]. In the case of a swarm, we can use cooperative localization as in [54], [55]. The typical literature uses beacons or visible landmarks; more interesting methods are information exchange to reduce errors in dead reckoning navigation, or bio-inspired localization [56]. In our paper, we do not perform absolute localization but only configuration, and in a totally new way, using only the distances and speed (or orientation angles) exchanged between the machines.

Robot localization and mapping are commonly related to cartography, combining science, technique and computation to build a trajectory map that can be used to correlate spatial information with the data collected by the sensors. Therefore, an autonomous robot should be able to construct (or use) a map and localize itself within it. The localization achieves not only the localization of the robot itself but also allows it to reach some objective: for example, a sensor positioned on the sea floor which gathers data or obtains a specific position in the sea (or a pool [57]) to monitor. A possible working method consists of distinguishing between different scenarios for the localization of the robot, i.e., a structured environmental scenario, or otherwise. We distinguish between the problems of localization for a single robot and for a swarm; the latter must solve the configuration problem, which we call 'relative localization'. However, the swarm also has the possibility of collecting information from many points of view to perform cooperative localization. Localization of a single entity is a problem relevant in understanding a metric of the space, possibly realized only by a sequence of reference points, and placing the entity in relation with the established metrics or with given reference points. Localization of a swarm requires different techniques that can be used with the different information that can be gained by putting them together. As an example, cooperative localization can be realized if the swarm recognizes a landmark by combining different pictures taken by different robot [38], [54], [58]. The problem can be considered as a system with many inputs (distances, speeds and orientation angles) and many outputs (the relative positions of the machines), which is therefore a MIMO system.

The localization concept does not necessarily coincide with the definition of a reference frame metric. For many animals, for example, localization may be related to different types of metrics, such as those linked to research sources for something that is critical to their life (or, for the swarm, a more important economy). Food for animals is the equivalent of a source of pollution in the mission for our swarm engaged in environment monitoring. They are both represented as "needs" of the swarm.

In this case, we refer to identification on a physical field with gradients of arbitrarily complex shapes; our position therefore refers to the smell, small or large in magnitude, of our relative reference parameter (i.e., oil presence). We can move towards the higher value of the field, the oil trail, in order to reach the source. In this circumstance, the geometric or geographic locations are meaningless for the purpose of the mission, but communication with a human supervisor requires reference to the geographical location.

The gradient localization is an example of not metric but topological localization; its realization is similar to the force field used in obstacle avoidance techniques.

Another example is the kinaesthetic sense of the human body; it works with sensitivity to a muscle extension that is detected by non-metric variables. The brain later relates these variables to position the body within the world around it, and thence to a geometric structure.

Of course, to perform this kind of localization related to a parameter, the swarm must be able to obtain the parameters' correct values using sensors, i.e., if we locate by smell then we must have a nose. It is important to note that the meaning of a map is very general. It is a multidimensional set of associated features (houses, streams, odours) that allow you to switch gradually from a certain place to another place via a route that is typically (but not necessarily) geographic.

A tracking system used by humans and animals is constructing a map of the environment by fixing in the memory a set of recognized items that are connected to each other. These kinds of items are very general and are not necessarily geographic in nature (although those are by far the most important type). Moreover, the human mind often does not work geographically. Rather, the mind associates the recognition of a number of characteristic features (houses, streams, odours) with other distinguishing features (other homes, streams, smells, plants) that have the characteristic of "proximity" with those that preceded them. The "map" thus becomes a multidimensional set of associations.

The number of dimensions of this approach is relative because it depends on many variables, some of which are difficult to measure. Moreover, for a human, this number is relevant only in relation to other needs (e.g., the time it takes to go from one place to another).

The geographic approach, supported by the geometric metrics, is a powerful tool that humans have developed and which allowed them to make huge progressions relative to those of other species, but is not essential to the success of a mission and is not the only factor.

In the case of a swarm or a non-rigid single machine, its localization relative to the other members of the swarm is also relevant. Since the shape of the swarm is very important for the mission, relative localization has the same importance as absolute localization; we can use the same conceptual categories defined by absolute location, which are, again, not necessary geometric. However, leaving the purely conceptual world and progressing to basic concepts, we may say that the swarm's location can ignore these issues and extend beyond the location of a geometric type. Now, we will show briefly some kinds of localization in metric form.

The easiest form of metric localization to understand is that using a Cartesian frame system, and in this contest the simplest form of localization is the open loop estimation, which has the means of estimating position based on expected results of motion commands. Therefore, no contribution from the sensor is required and no feedback is calculated.

Without an external reference, such as acoustic beacons at known positions, the vehicle has to rely on proprioceptive information obtained through a compass, a Doppler Velocity Logger (DVL) or an Inertial Navigation System (INS). To this end, one of the commonest methods totally internal to the robot family is the dead reckoning, which is the most common method after the open loop.

The information that the robot gathers can be divided into two kinds: those from idiothetic and allothetic sources. That information involves internal and external sensor source data; for example, if a robot is counting the number of wheel turns in order to calculate space, this is classed as an internal source.

The allothetic source corresponds to the sensors of the robot, such as a camera, microphone, laser or sonar. A typical problem of this method is "perceptual aliasing"; this means that two different places can be perceived as the same. For example, it may be impossible to determine your location in a building because all corridors look the same; this is sometimes true for humans dependent solely on visual information.

The most commonly used dead reckoning sensor is an INS. An INS measures the linear acceleration and the angular velocity of the vehicle using three accelerometers and three gyroscopes. Typical underwater external sensors used to correct accumulated errors from the integration of the INS measurements, are Doppler Velocity Log Sensors (DVL), Ultra Short Baseline (USBL) and Differential Global Position Systems (DGPS/GPS); the latter is only used when the vehicle is operating in shallow waters which it can leave in order to fix (and eventually communicate) the position.

Independent of the quality of the sensors used, the errors in the position estimations based on dead-reckoning information grows without an upper limit. Typical navigation errors include the distance travelled by vehicles travelling within a hundred metres of the sea. Lower error rates can be obtained by using large and expensive INS systems, but error rates for vehicles relying only on a compass and a speed estimate can be higher than 10%, after 100 metres. The errors can be reset if the AUV comes to the surface by GPS, but this is sometimes impossible (when under ice, for example) or undesirable (during a security operation) [7]. The use of beacons to form a Long Baseline (LBL) array limits the operation area to a few square kilometres and requires a substantial deployment effort before operations to pose the beacons, especially in deep water. This reduces the advantages of AUV and requires use of an expensive ship to support the operation.

Other methods employ the use of a landmark. If we use external references, like humans, we have to deal with their definitions and position them on a map (metric or otherwise). Any kind of landmark is subject to classification based on its attributes. Moreover, the identification of a landmark often suffers from ambiguity, owing to the multiple solutions of the associate equations describing it.

A set of features' location estimates can be thought of as a map. The challenge is to combine INS/dead-reckoning and other information with sensor observations of features to build a map, either locally or globally referenced.

A more modern method consists in matching measurements of one or more geophysical properties, such as bathymetry, gravity or magnetic fields, to an a priori environment map [59], [60], [45], [61], [62]. If there is sufficient spatial variation in the parameters being measured, there is the potential to reduce navigation uncertainty. A turtle's migration, for example, is monitored by magnetometers measuring the earth's magnetic field variations. However, these techniques often require an a priori map of the environment that is not available. The marine turtle's migration is monitored by a magnetometer with three axes, but the resolution of this system is 35 nautical miles [63], [64] .

It seems to be a good idea to divide the localization problem for a swarm into three tasks:

1.      Absolute localization (AL), has the meaning of localization of one member of the swarm with respect to a fixed reference system.

2.      Relative localization (RL) of a swarm's member with respect to the other members' configuration.

3.      Relative localization of a member with respect to other swarm members, but only its neighbours; we call this immediate relative localization (IRL), and its meaning will be elucidated later.

Different methodologies are required to solve the three tasks. In agreement with the swarm philosophy, each element must be able, if connected with the others, to perform the localization job. Pay attention to this last statement. We do not mean that each element must do all jobs; often, there is no need to know the positions of all machines. However, we must be able to use all the internal data, the external data communicated by the other elements, and the external data measured by the robot, including those deduced by observation of the environment.

The usefulness of the third categorization (IRL) was designed to obtain a rapid response to variations in the environment of the entire swarm. If the system had to attain thorough knowledge of its internal structure before deciding on the type of response of each unit of the swarm, the reaction times would be far slower than the dynamics of environmental phenomena, and the system would have a low chance of succeeding in its mission. The recognition mechanism used by a school of fish that allows its members to rapidly change the position of the entire school just by looking the movement of their neighbours is well known. In principle, IRL is similarly more easily achieved using simple signals (which are very fast and can be transmitted quickly) based on a physically fast transient mechanism, and possibly using a rapid propagation of fields such as the optical field. A livery, like fishes, is very useful to this aim. It does not need heavy image processing but only, for example, a measurement of reflected light. Later, if necessary, the swarm can compute the relative position of the whole system.

## 6. Needs of scenarios

To address the localization problem of our swarm, we define three types of operative scenario:

1.      The swarm is in the open water approaching the objective; eventually, monitoring is allowed. This should be the simplest localization situation; one element of the swarm is on the surface and fixes the position by GPS. The relative position of the other members is measured by other methods. Swarm shape is dependent on the task.

2.      The swarm is operating very close to the objective. The shape of the swarm is able to maximize the area covered close to the objective. Landmarks of the objective are useful for localization.

3.      The swarm is operating in a structured environment such as a harbour. This is the most difficult task, because the aim of the mission (surveillance, pollution monitoring, etc.) that determines the swarm's movements and shape must take into account the constraints of the environment. In fact, in such conditions, some shapes are not possible. As an example, in a harbour it might be necessary to minimize the volume occupied by the swarm to avoid collision with the ship or other objects, and the use of an acoustic beacon could be convenient.

In these three different scenarios, the strategies of movement and localization might be completely different.

## 7. Equipment

Equipment to perform these solutions in the HARNESS project is divided between "basic requirements", which is the minimum instrumentation we anticipate having in the single machine, and "desired requirements", for enhanced instrumentation and better performance. Quantitative considerations for each apparatus are not reported here due to lack of space.

The basic equipment of all the machines is composed of Network communication, GPS, Depth meter, Inclinometer, Compass, Flash, Photodiode, Web camera, and Livery on the vessel's surface. All this equipment is cheap and readily available.

The network is a requirement that exists not only for communication but must also be used for data exchange; we are interested in its use in sonar ranging for RL. Of course, the snapshot of configuration suffering a delay, presumably from durations of tenths of a second to second, should be sent for correction together with the estimated robot speed. The network should be able to shift the working frequency from 100 to 1000 KHz (at least two frequencies). This number is calculated by consideration of data rate and the use of the net as an emergency ping or localization signal. The distance we want to cover (a maximum of 50 metres) and the data rate should be between 10 and 100 Kbytes/sec.

All the available data will be fused and weighted with all the data coming from other instruments to enhance precision in the localization task. More than one algorithm is required; for example, a more complex one that uses all the available data and a faster one using only a subset of data, depending on the operating conditions and on the kind of localization required.

GPS is used for AL when a scout (a single robot that has been given this task) is on the surface. At commercial depth meter must be present for AL in the z dimension.

A couple of inclinometers are useful to measure the angle of position (yaw) with respect to the bottom of the sea. Another degree of freedom can be eliminated by using a compass, which must be posed carefully in case magnetic disturbances are present.

A photodiode is useful for obtaining the flash lamp sequence for light communication; as an example, a codified sequence could send the heading of the machine to close neighbours to change them. Therefore, we can transmit simple codified messages by flash to transmit a sequence; working on colour or flash times is more complicated; remember we are using a cheaper flash unit. In some cases, we can obtain an optical modem. This is a much cheaper and lighter instrument, but one that does not give information on the position of the light source. Works in progress are considering the use of flash lamps to calculate distances, in spite of the different water transparencies that modify transmission parameters.

A web camera with a flash lamp must be used for image recording. Moreover, by using omnidirectional vision it is possible — with flashes synchronized or triggered by the first flash and using cumulative vision for a few seconds — to get qualitative information on the density of machines and their neighbours. Of course, this is not metric information, but a single machine can obtain the information if it is far to the left (for example) of the swarm. Moreover, flashes could be useful to rescue a single machine experiencing difficulties, together with a switch of the network towards a lower frequency, working to increase the range distance.

The camera can also be used in the livery vision for IRL for fast reaction movements, but it requires a computational job that must be simplified, because image analysis is much too demanding. This last job could be performed as a batch as in Simultaneous Localization and Mapping (SLAM), distributing the computational cost on a parallelized machine (the swarm itself, if the network band-pass is adequate).

In the future, the machines can also be equipped with some more instruments, as described in the following paragraphs.

Some other pieces of equipment, such as an acoustic pinger; an LBL USBL device; DVL and AHRS electromagnetic devices; and gradient localization devices can increase the performance of the system. A harbour could be equipped, by USBL, at a reasonably low cost with respect to the normal costs of surveillance. A DVL device can be mounted to integrate its data with a cheaper AHRS device (like XSense, for example) for AL and RL (we avoid IMU, owing to the cost). Magnetic (and electrical) methods require further discussion. A quantitative measure of the absorption of electromagnetic waves in the water led to the realization that the radio frequency modem cannot be used in seawater. However, some new electromagnetic underwater modems offer promise for the topic under investigation; electromagnetic transmission in sea water is making progress with respect to that of last year [65], [30]. They saw an improvement in their performance leading to the development of commercial products [66], [67]: these new devices need to be investigated, since their dimensions seem to be too large for the HARNESS project. Electric and static magnetic fields also are under investigation. The machines could be equipped with a strong electromagnet. The advantage of a static magnetic field is in its possibility of transmitting some information to the other machines, like in a flash sequence. Moreover, the hope is that, unlike in the light source, it might be possible to obtain some quantitative information on the RL by the magnetic field vectors. An attempt to localize RFID using a magnetic field has been performed in air [68]. We started with [3], considering the Earth's magnetic intensity field. Its scalar value is about 20 MicroTesla at the equator and 70 MicroTesla at the poles. We can consider these as constant values in our area of operation, with some exceptions. We can generate a known perturbation in magnetic Earth field to get information on the perturbation position, (distance or direction, among others) from these numbers. We calculated in [3] that a cheap magnetometer is able to do this (also taking into account the natural anomalies of the magnetic Earth). So far, we should be able to detect the spike in the magnetic field (we have calculated the values for some conditions) superimposed onto the Earth's field that we generated in the sea. No information was obtained from the indiced transient of the field but we measured a change in the magnetic Earth fields. We now have two opportunities; one is a very slow modulation of the field (to reduce attenuation of the field) carrying some codified information, like a flash lamp. The second is to make an attempt to calculate the position of the field generator. This has been done for two objects in open space. It is difficult to do this for a multisystem in the sea but we are investigating the possibilities.

## 8. The localization problem: our first proposal

Now we focus our attention on the relative localization problem, i.e., how to determine the swarm configuration from the signals that the robots exchange between themselves. Absolute georeferenced localization will be available only when one of the members emerges to fix its GPS position, or when some georeferenced map position or landmark is available in the water.

Each machine is characterized by six degrees of freedom, but two of them (depth and heading) are very easy to measure, using a depth meter and a compass. If the machine has cylindrical symmetry, one is uninfluenced. Considering the yaw is not important (our images of the torpedo are always quite parallel to the ground, with the exception of a few acceleration seconds); we understand that the real difficulties arise from the coordinates x-y of the centre of mass. The x-y plane is referred to as parallel to the bottom of the sea.

We shall consider two possibilities: one when the robots have a speed meter and a second when they do not. Two different methods are presented to achieve the same aim: which is the geometric configuration of the swarm in space.

To obtain the complete configuration, it is enough to choose one member as the axes' origin and obtain the relative positions in space of three or (better) four or more of the other members as a constellation. At a later date, any other member will be able to localize from its distance from the constellation, using algorithms commonly used in GPS calculations [69], [70]; in this manner, the initial members are used like a constellation of satellites. Therefore, we now need to localize some members, with respect to one [69]. Once we obtain the constellation, localizing the whole swarm by distance is an easy and knowable task.

In some fortunate circumstances we might have some vessels on the surface (which can be substituted by little boats), and we could take advantage of a high band-pass using a laser communication system (if one is available) and an easy transmission vertical channel. In fact, the collimation problem between vessels typical of laser transmission can, in this case, be partially avoided (the solid angle has been well selected).

If we do not have the surface vessels, we can calculate the Relative Localization (i.e., the configuration of the swarm) by solving the distance equations, and create a constellation by using a fairly similar trilateration problem.

Now we can see how to obtain the coordinates of the vessels that we shall use, and their constellation.

What kinds of signal are exchanged between the robots? Remember that communication is a problem in an underwater environment, so we must achieve results with a minimum of data exchange. This means that all communication between the robots must be used to gain information on their positions; we decide that each communication must contain at least the identity of the transmitter and the starting time of transmission. This is not enough for our purposes, so we must include in the message something more, but with a minimum increase of the bytes to be transmitted.

In a previous work [3], we showed that configuration can be obtained using the following elements transmitted in communications: 1) ID of the vessel; 2) time; 3) speed; 4) its neighbours' data for the three previous elements. This is called the "heartbeat" of the swarm, and these data must be available not only for the member itself but also for its neighbours.

Heartbeat signal = {ID, time, speed, neighbours' distances}

Therefore, each robot transmits these data, probably together with some other communication. We are considering, here, the x-axis oriented in the northern direction.

From the time of flying, we obtain from the acoustic signals the relative distance between the robots. At this point, we face a trilateration problem but, unlike in the standard problem, we do not know the position of the beacons so it is a very difficult calculation. From an algebraic point of view, this problem is classified as a non-polynomial hard problem. It has some similarities with the problem of determining the conformation of the proteins with from the molecular distances obtained by NMR experiments [71], [72], [73], [74], [75]; in our case, however, we are sensitive to some symmetrical conditions that make the calculation more difficult. In fact, starting from the distances between the robots we obtain many possible solutions, owing to the high degree of the non-linear equations system. We have to use some more information to choose the only right one, corresponding to the real situation.

We have built a simulator, using the Mathematica software by Wolfram Research, able to manage this task. Consider Figure 6, depicting some grey robots; their positions are unknown to themselves, so one of them (the green one) decides to be the axes' origin and tries to build the swarm configuration. Later, it receives the information relative to three other robots (six distances and three speeds). We now have our equations system.

$$d_{i,j} = \sqrt{(x_i - x_j)^2 + (y_i - y_j)^2 + (z_i - z_j)^2}$$

With i,j = 1,2,3,4.

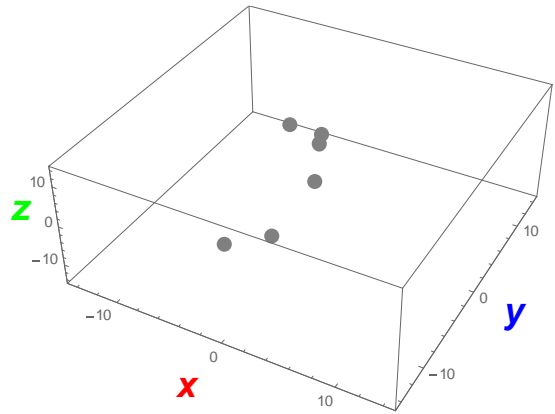

**Figure 4**. The initial configuration, in which each robot ignores the existence of the others. When one member receives a signal to become green, it decides to be the axes' origin and starts to calculate the swarm's configuration.

We have solved the system and have eight possible solutions as shown in Figure 7, where the blue robots are the possible candidates. Note that we represent seven machines instead of six, owing to the multiplicity of the solution.

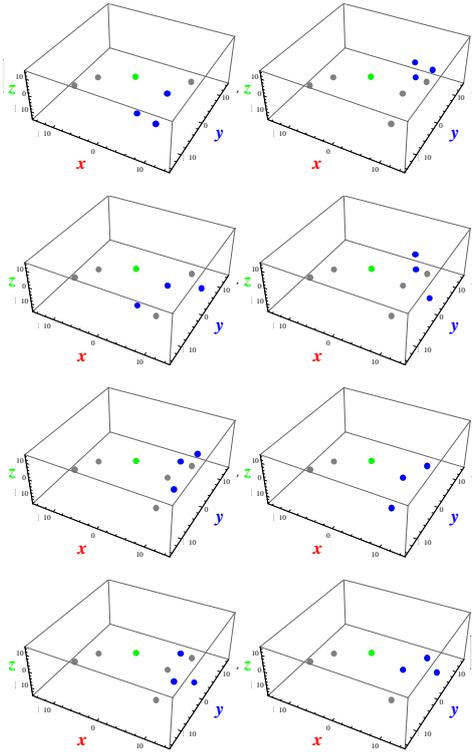

**Figure 5.** The first eight possible configurations of six vessels. Note the presence of seven points, except one owing to the degeneracy of the solution. Grey vessels are unknown in their existence, the green vessel is chosen as the axis origin and blue vessels send a heartbeat signal to the green one.

Now we cannot go any further, and in order to select the unique correct solution we proceed as follows.

We have multiple possible solutions and need some more information in order to select between them. A flash, for example, to determine the direction from which the signal arrived could be useful information. A cleverer system is to use a any (but known) movement of the swarm and to repeat the calculation for the multiple solutions corresponding to the newly possible configuration. Repeating the calculation after the displacement, we obtain a new set of eight possible solutions. Using the preceding configuration and applying a transformation from the known movement, we now obtain only one configuration matching the old and the new. In other words, if we have the speed of this set, we repeat the measurements after a known movement. At this point, we have another eight possible solutions but only one of them is compatible with the displacement, using coordinates changes given by the known movement. The final solution is visible in Figure 8, represented by red dots.

This is because each solution must conform to:
$$x_{ik} = v_i * t$$
and
$$y_{ik} = v_i * t$$
where k is the k-ma possible solution (k=1,8) and $v_i$ is the known speed of the i-ma machine.

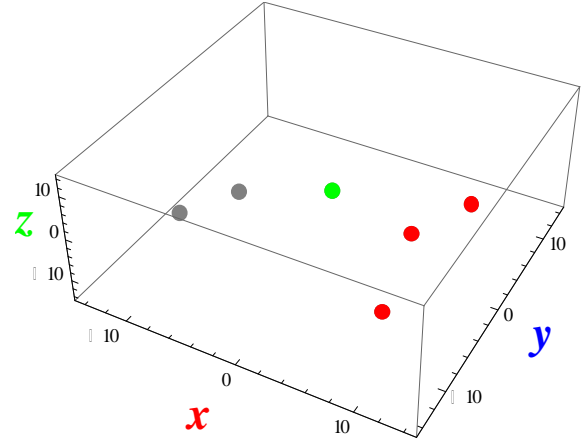

**Figure 6.** The final solution obtained after a known movement and coordinate transformation between couples of eight solutions.

Note that if the movement is a uniform translation, we obtain no more information and the problem cannot be solved; luckily, in the real environment, this case never appeared. So far, we have our constellation.

Reassuming our simulation algorithm, we generate the position of the vessels and proceed in two steps

1) Building a constellation reference.
2) Classical trilateration calculation using a GPS system (non-linear least squares, circle intersections, etc.)

One member receives a "heartbeat" broadcasting signal (a low frequency signal that each vessel transmits) and is chosen as the coordinate origin (the green vessel). The signal contains ID, time, its neighbours' distances and the last estimated movement vector.

Then this vessel (any of the vessels can do this job) calculates the distance of the closest surrounding vessels (represented by blue) and the distances between them, whereas the others (in grey) will be ignored and do not exist at this time. This can be done using whatever signal is received from the vessels.

Solving the equation system of a similar trilateration problem (but with stations' coordinates unknown), we obtain the eight possible configurations of the three vessels (the blue dots); this is because the problem has multiple solutions owing to its symmetrical geometry, knowing only distances. The possible solutions are shown in Figure 7. The grey robots are still unknown (do not exist) and the blue robots are candidates for the configuration. Note that only one of the eight is correct, but that which one is correct is unknown at this time. We have multiple possible solutions and in order to select between them we need some more information. The reason for this is that, as in Figure 7, there are some blue ghost robots that do not match the corresponding grey ones, and seven instead of six vessels are visible in Figure 7. Now it is possible repeat the process for the other vessels. Repeating the calculation after the known movement solves the configuration problem, as can be seen in Figure 8, and we have our constellation.

The flow chart is:

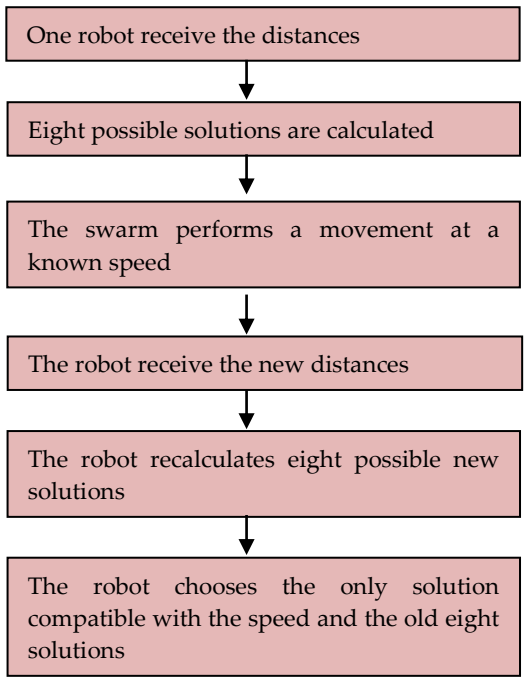

Many improvement of this method can be applied.

As an example, to overcome signal delay problems (some vessels could stray too far from the original coordinates), another vessel can be taken as the origin coordinates and the found configurations can later be matched together, taking into account the different time measurements and using sped to interpolate positions. We can use these as a base constellation for the others, such as GPS satellites. So far, for these other robots, we only need the distances between them and no other movements, because the constellation is known.

Multiple combinations of the method, to minimize errors, are possible. More than one constellation is possible, adding new members and repeating calculations. The coordinates of a new member can be calculated using constellation distances or another known movement.

Later, the constellations can be varied and updated with new data, repeating the process for other vessels. This work can be done in different volumes of the swarm and matched together later; this entails the clustering of the swarm if it is too large to be managed in a reasonable time.

So far, we have eliminated the multiplicity of the solution of a trilateration problem by a generic, but known, movement of the swarm. We need, therefore, two heartbeats in two different but preferably close time periods. This method uses every signal exchanged between the swarm elements.

Of course, it is necessary to be careful of noise in the distance measurements, making an acceptance protocol of the data necessary. An optimization proceeding, instead of solving an exact algebraic equations system, should also be used to avoid equations systems without solutions; this is the subject of the second method we propose.

## 9. The localization problem: a better solution

Now we want something both simpler and cleverer. Because the vessels are very cheap, we might not have any information about speed, which, moreover, is often affected by many errors; therefore, we use an optimization procedure with constraint. However, we have a depth meter (so the z coordinate is known, see Fig. 9, where the tracks of three robots are plotted) and a cheap compass that give us the orientation of the robots. The problem is, therefore, bidimensional, considering the projections of the trajectories on the x-y plane. Our second simulation algorithm can demonstrate that these, more cheaply obtained, data are enough to solve the degeneration of the solutions in one or more steps of movement, with some differences with respect to the old method. Our aim is always to obtain the satellites' constellation, comprising at least three machines, to use to build the configuration of the entire swarm.

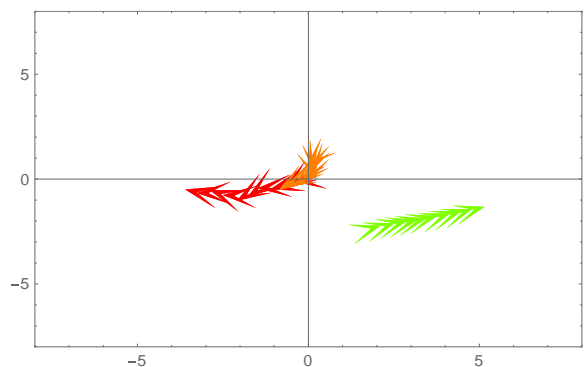

**Figure 7.** Three robots and their movements at different points of time.

So far, in this case, our heartbeat signal will be = {ID, time, orientation, depth, and the neighbours' information for those factors}. We are also considering north as the direction of the x-axis.

The distances between the vessels are always calculated by the time of flight of the signal, using all the acceptance and errors procedures, to take into account the fading, reflections and the many errors affecting the measurements.

It must be considered that by using orientation information, we are working with an inequality condition on the coordinate in a different time. This means that many compatible conditions exist, and that a single step cannot be enough to solve the degeneracy of the solutions in the equation's system distances. Three or more steps of motion might be necessary (in the simulation, we found up to seven) to resolve the degeneracy of the equations system.

The advantage of this cheaper algorithm lies in the very small amount of data that must be transmitted in each communication. Moreover, we can recover the speed of the robots and all the tracks once we have calculated the configurations. This is a significant result because a speed measurement in water is always a hard one to take. Note that at the beginning, we are only using three robots; if additional information about a fourth or more robots is available, it can be used to resolve the degeneracy more quickly. Typically, a situation with only three robots cannot be solved quickly.

Unfortunately, the ideal situation studied is unrealistic as the data are sure to be affected by many errors. This method only works in our simulation and might cause problems in an experimental campaign. We have tried inserting random errors in the simulation for the distance measurements and this led fairly quickly to equations systems without solutions. The use of inequalities, together with the errors affecting distance measurements, forced us to abandon the possibility of solving an equations system.

So far, we developed a new algorithm using a minimization procedure of one or more objective functions, which works much better with inexact data and the many constrained conditions expressed in inequations form. The advantages are that any new variables, together with their constraints, can be added very easily to the algorithm. The presence of a new machine can also be added very easily. The constraints on the variables lead us to search for the solutions. So far, we have the same distance equation as before:

$$d_{i,j} = \sqrt{(x_i - x_j)^2 + (y_i - y_j)^2 + (z_i - z_j)^2}$$

With i,j = 1,2,3.

The constraints are of three kinds.

The first is given by the orientation angle: if the cosine of the angle at time $t_h$ of the robot i is positive, then the x

coordinate of the same robot at time $t_{h+1}$ is greater than the preceding one, and so on. Therefore:

If $\cos \alpha_i (t_h) > 0$ Then $x_i(t_{h+1}) > x_i(t_h)$

For the sine of the angles, there are similar relationships. We are assuming that the orientations do not change much in the time period considered; this curtails the number of possible solutions.

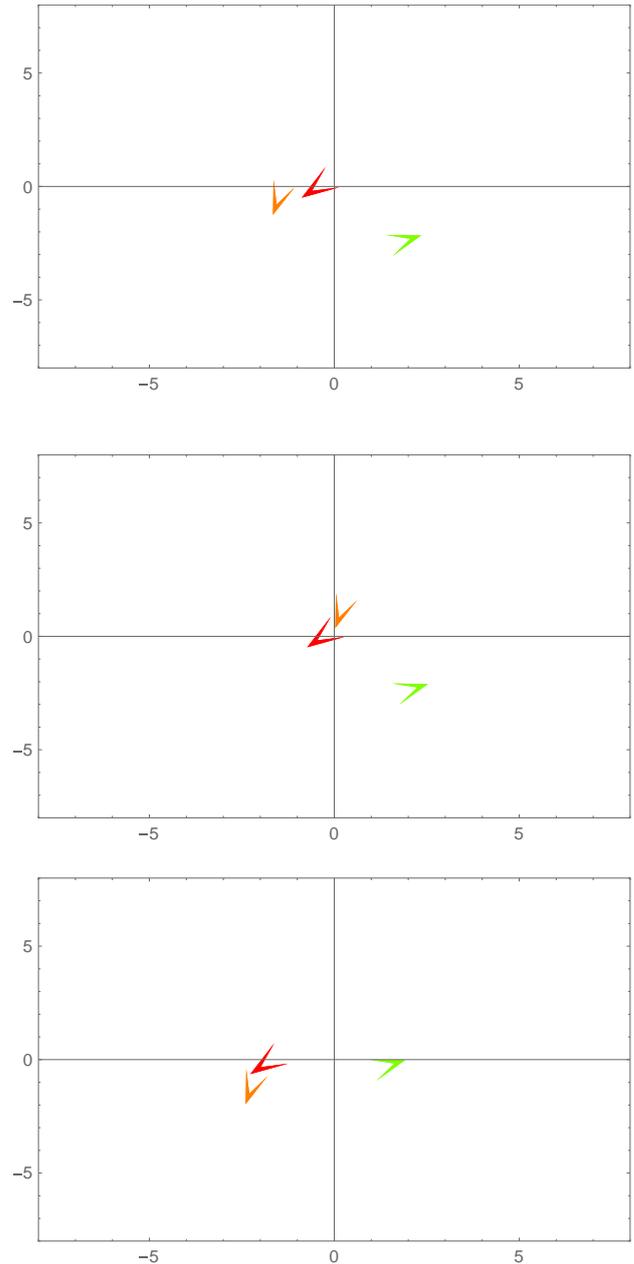

**Figure 8.** Example of multiplicity of the solution after two steps. All the solutions are possible but only one (the second) is the real one; we need one or more motion steps to identify it.

The second kind is regarding the distances; the $x_i(t_h)$ (coordinate of robot i at time h) is linked to the coordinates of the other robots, at the same time, by the distances between them. Note that if I have no distances to input, that is no problem for the algorithm, it is just that some constraints are missing; the result can be obtained all the same:

$$x_i(t_h) < x_j(t_h) + \text{dist}(P_i(t_h), P_j(t_h))$$

and

$$x_i(t_h) > x_j(t_h) - \text{dist}(P_i(t_h), P_j(t_h))$$

The third kind of constraint comes from the maximum speed; we do not know the robot's speed but, without a strong current, we can affirm that variation in the x coordinate within a known time interval cannot be greater than the maximum speed multiplied by the time interval:

$$x_i(t_{h+1}) > x_i(t_h) - v_{max} * (t_{h+1} - t_h)$$

and

$$x_i(t_{h+1}) < x_i(t_h) + v_{max} * (t_{h+1} - t_h)$$

Every new constraint is well accepted and can be inserted into the algorithm. The algorithm also has a tolerable acceptance for constraints, so contradictory constraints resulting from errors in measurements, can be matched. It has been developed using the Mathematica software by Wolfram Research, but can also be implemented in C language on board the vessels.

Therefore, we used a minimization procedure for one or more objective functions. The base objective function to be optimized is the sum of the square of the distance formula less the measured distances, together with the angular coefficients' formula less the measured one. This should be close to zero (equal to zero, when the equations system is used).

It is well known that there are many methods in optimization; if the system or the conditions on the variables are non-linear, there are no sure methods of obtaining a global minimum of the objective function. Therefore, we have also tried many optimization methods in the multivariate mode, using more than one objective function. An example of a multi-objective function can be obtained if we separate the first function (the sum of the square of the distances) into two parts.

In Fig. 9, the trajectories of three robots over different periods (10 time steps) are shown. When we tried to solve the configuration problem, see Fig. 10, after two steps of periods we get three possible solutions; only one is the real one, but we do not know which and have no way of choosing. We need another step in order to solve and obtain a unique solution. Note that a swarm composed of only three robots is more subject to a multiplicity of solutions, owing to the large number of possible symmetries.

In some particular conditions, we can use some tricks to obtain our solution. As an example, a nice system to obtain a global minimum is achieved taking into account two factors. As first, we are working within a circle of 50 metres' radius (the maximum distance between two robots). Moreover, we are interested in a precision of 0.1 metres. Therefore, we can discretize our space and work with integers numbers. This means that the x coordinates, ranging over 100 metres, can assume 1,000 different values. The ones on the y-axis depend on their distance from the origin. The number of possibilities increases very quickly; two robots, more that the axes' origin has 1,000 x 1,000 x possibilities of $x_2$ and $x_3$ (x coordinate of Robots Two and Three, respectively). The constraints, however, prune the number of those possibilities very quickly; they become of the order of 1,000, taking into account the fact that Robot Three must be at a fixed distance between the first robot (the origin) and the second. Therefore, it is possible to realize a brute force algorithm that can calculate the global minimum of the objective function (Mathematica examines ten million possibilities in one second).

Another trick, introducing a few errors into the solutions, could be the following. If we have the values for $x_i(t_h)$ (x coordinate of machine i at time h), then it will be easy to solve the system equations (as in the old method) or the objective function in order to obtain the configuration of the swarm. Unfortunately, we do not have those values. However, if two robots are very close each other we should consider the $x_i(t_h)$ coordinate equal to distance, and the calculation then becomes very quick to solve. We have calculated that the errors we introduce by this approximation are quite equal to the errors between the real $x_i(t_h)$ coordinate and the distance. This error affected all the other calculated coordinates and does not grow with the increase in time or numbers of steps. Therefore, in some cases we can use this approximation to obtain a quick result.

Another possibility is to use the flocking rules [76], [77], [78], [79], [80], [81], [82], [83], governing the behaviour of the single elements of the swarm, to curtail the number of less probable configurations.

We also have introduced some random errors in the distance measurements; the minimization method has the advantage that we obtain the same configuration.

The flow chart describing the procedure is the following:

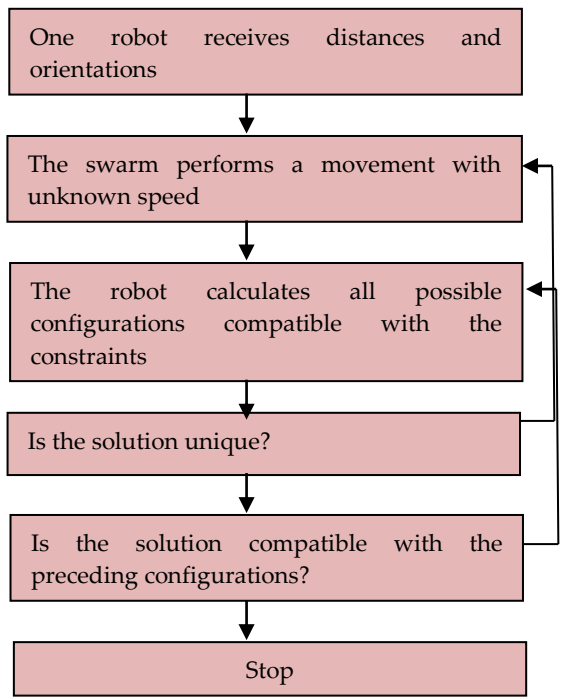

| One robot receives distances and orientations |
| :--- |

| The swarm performs a movement with unknown speed |

| The robot calculates all possible configurations compatible with the constraints |

| Is the solution unique? |

| Is the solution compatible with the preceding configurations? |

| Stop |

## 10. Open questions and future work

Some considerations must be underlined and some points must be clarified regarding the work presented in this paper, as work remains to be done.

In particular, we noted that the convergence speed of the algorithm depends on the initial configuration. So far, more than one step of motion is necessary to solve the swarm configuration. The simulation gives results in which the more similar the initial conditions between robots (orientation and unknown speed), the more steps are needed; though if you get more than one solution, they are often very close to each other. How many steps of movement does one need to obtain a unique solution? It depends on the initial positions. How do the initial positions influence the number of steps? These are some questions we are investigating to obtain a quantitative treatment. Therefore, we are working on a better mathematical treatment to determine how many steps one needs and how those steps depend on the initial configuration.

Sometimes, the minimization method did not converge in a reasonable time (generally, a few seconds) or it stopped on a local minimum of the objective function; moreover, no solution was available by the old method of solving the system equations. We still need to understand what is locking the procedure. Once again, this requires a more refined mathematical treatment. In any case, a pruned brute-force algorithm seems to be promising.

The robots do not emit their signals at the same time: interpolation needs to occur. Considering the slow speed of the machines, this should be not a problem but it must nevertheless be quantified.

We have to put together configurations calculated in different times and places. This means that we have put together all the possible configurations, if any one element of the swarm performs the same calculations considering itself as the axes' origin; this reduces the errors in coordinate calculations.

Finally, we have to build the final configuration of the whole swarm, putting all the pieces together.

## 11. Conclusion

In this work, we have explained how the HARNESS project could be useful for coastal monitoring, surveillance and many other purposes; we propose deployment of a swarm of underwater, cooperating robots. The advantages lie in the economy of the method, the parallelization of the task and the robustness of the system. The disadvantage lies in the major difficulty of controlling the swarm, owing to the presence of a new layer named the "Swarm control" which has different rules from the individual machine control. Many difficulties remain to be studied, especially in the communication between swarm elements, owing to the unfriendly environment that limits the communication channel if different methods are used together.

In this paper, we focused our attention on the localization problem; we have generalized and enhanced the work presented in the Bio Inspired Robotics Conference in May 2014. Acquiring knowledge of the swarm configuration problem, which is only one aspect of the general localization problem, is a very important task; it has been solved using very few pieces of data exchanged between the elements of the swarm to minimize the use of bandpass. We have developed a simulation algorithm concerning the motion of a swarm and the signal exchanged between the individual robot machines, which is able to build the configuration of the system. Two different methods have been implemented. The principal difference is that in the first we need a speed meter on board, while in the second we do not. Moreover, the second method is less sensitive to the errors affecting the measurements, using an optimization procedure.

Some aspects of the convergence of the algorithm must still be investigated, together with a better mathematical treatment of the theory, but some suggestions have been given in this paper.

The work is in progress in our laboratory, with an experimental campaign in the Bracciano Lake.

## 12. Acknowledgements

## References

[1] R. dell' Erba and C. Moriconi, HARNESS: a robotic swarm for environmental surveillance, in *6th IARP Workshop on Risky Interventions and Environmental Surveillance (RISE)*, Warsaw, Poland, 2012.

[2] R. dell' Erba and C. Moriconi, HARNESS: a robotic swarm for harbour security, presented at the International Workshop Port and Regional Maritime Security Symposium, Lerici, Italy, 2012.

[3] C. Moriconi and R. dell' Erba, The localization problem for HARNESS: a multipurpose robotic swarm, in *SENSORCOMM 2012, The Sixth International Conference on Sensor Technologies and Applications*, 2012, pp. 327–333.

[4] R. dell'Erba and C. Moriconi, Bio-inspired robotics. Available at: http://www.enea.it/it/produzione-scientifica/edizioni-enea/2014/bio-inspirede-robotics-proceedings. [Accessed: 15 December 2014].

[5] J. J. Leonard, A. A. Bennett, C. M. Smith and H. Feder, Autonomous underwater vehicle navigation, in *IEEE ICRA Workshop on Navigation of Outdoor Autonomous Vehicles*, 1998.

[6] B. Anderson and J. Crowell, Workhorse AUV-a cost-sensible new autonomous underwater vehicle for surveys/soundings, search & rescue, and research, *Proc. MTSIEEE OCEANS 2005*, pp. 1228–1233.

[7] H. Brown, A. Kim and R. Eustice, An overview of autonomous underwater vehicle research and testbed at PeRL , *Mar. Technol. Soc. J.*, 2009vol. 43, pp. 33–47,.

[8] M. A. Joordens, Design of a low cost underwater robotic research platform, in *IEEE International Conference on System of Systems Engineering, 2008*, pp. 1–6.

[9] V. Kopman, N. Cavaliere and M. Porfiri, MASUV-1: A miniature underwater vehicle with multidirectional thrust vectoring for safe animal interactions , *IEEEASME Trans. Mechatron.*, ,vol. 17, n. 3, pp. 563–571,.

[10] J. Yuh, Design and control of autonomous underwater robots: a survey, *Autonomous Robot*, pp. 7–24, 2000.

[11] G. Beni, From swarm intelligence to swarm robotics , *Swarm Robot.*, pp. 1–9, 2005.

[12] G. K. Venayagamoorthy, L. L. Grant and S. Doctor, Collective robotic search using hybrid techniques: fuzzy logic and swarm intelligence inspired by nature, *Eng. Appl. Artif. Intell.*, vol. 22, n. 3, pp. 431–441, 2009.

[13] E. Bonabeau, M. Dorigo and G. Theraulaz, *Swarm intelligence from natural to artificial isystems*. New York: Oxford University Press, 1999.

[14] K. M. Passino, T. D. Seeley and P. K. Visscher, Swarm cognition in honey bees, *Behav. Ecol. Sociobiol.*, vol. 62, n. 3, pp. 401–414, 2007.

[15] S. Janson, M. Middendorf and M. Beekman, Honeybee swarms: how do scouts guide a swarm of uninformed bees?, *Anim. Behav.*, vol. 70, n. 2, pp. 349–358, 2005.

[16] A. J. C. Sharkey, Robots, insects and swarm intelligence, *Artif. Intell. Rev.*, vol. 26, n. 4, pp. 255–268, 2007.

[17] C. M. Topaz, A. J. Bernoff, S. Logan and W. Toolson, A model for rolling swarms of locusts, *Eur. Phys. J. Spec. Top.*, vol. 157, n. 1, pp. 93–109, 2008.

[18] J. Bachrach, J. Beal and J. McLurkin, Composable continuous-space programs for robotic swarms, *Neural Comput. Appl.*, vol. 19, n. 6, pp. 825–847, 2010.

[19] S. Chiesa, S. Taraglio, S. Pagnottelli and P. Valigi, Flocking approach to spatial configuration control in underwater swarms, ICINCO 2012 - Proceedings of the 9th International Conference on Informatics in Control, Automation and Robotics, Rome, Italy, 28 - 31 July, 2012, vol. 1, pp. 313–316.

[20] M. Dorigo, V. Trianni, E. Sahin, R. Groß, T. H. Labella, G. Baldassarre, S. Nolfi, J-L. Deneubourg, F. Mondada and D. Floreano, Evolving self-organizing behaviors for a swarm-bot, *Auton. Robots*, vol. 17, n. 2–3, pp. 223–245, 2004.

[21] F. Ducatelle, G. Di Caro and L. Gambardella, Robot navigation in a networked swarm, *Intell. Robot. Appl.*, pp. 275–285, 2008.

[22] S. Kalantar and U. R. Zimmer, Distributed shape control of homogeneous swarms of autonomous underwater vehicles, *Auton. Robots*, vol. 22, n. 1, pp. 37–53, 2006.

[23] K. Lerman, A. Martinoli and A. Galstyan, A review of probabilistic macroscopic models for swarm robotic systems, in *Swarm Robotics*, Springer, 2005, pp. 143–152.

[24] H. Liu, A. Abraham and M. Clerc, Chaotic dynamic characteristics in swarm intelligence, *Appl. Soft Comput.*, vol. 7, n. 3, pp. 1019–1026, 2007.

[25] Liu Bo, Swarm dynamics of a group of mobile autonomous agents, Vol. 22, No. 1 (2005) pp.254-258 *Chin. Phys. Lett.*

[26] D. Mirza and C. Schurgers, Collaborative localization for fleets of underwater drifters, *Oceans 2007*, pp. 1–6, 2007.

[27] K. R. Baghaei and A. Agah, Task allocation methodologies for multi-robot systems, *Proceedings of the IEEE Sponsored Conference on Computational Intelligence, Control And Computer Vision In Robotics & Automation,* March 10-11, 2008, NIT Rourkela, P 99- 106, 2002.

[28] C. Xian-yi, L. Shu-qin and De-shen X., Study of self-organization model of multiple mobile robot, International Journal of Advanced Robotic Systems, Volume 2, Number 3 (2005).

[29] Mark Rhodes, Electromagnetic Propagation through the Water Column, Available at: www.wirelessfibre.co.uk, [Accessed: 20 September 2013].

[30] Mark Rhodes, *Underwater Electromagnetic Propagation* — Archive — Hydro International. December 2006, vol. 10, Available at: http://www.hydro-international.com/issues/articles/id697-Underwater_Electromagnetic_Propagation.html. [Accessed: 26 October 2010].

[31] 6.013 — Electromagnetic Fields and Energy. Available at: http://web.mit.edu/6.013_book/www/book.html. [Accessed: 20 September 2013].

[32] J. Rice, SeaWeb acoustic communication and navigation networks, in *Proceedings of the International Conference on Underwater Acoustic Measurements: Technologies and Results,* Heraklion, Crete, Greece, 28th June – 1st July 2005.

[33] M. Dunbabin, I. Vasilescu, P. Corke and D. Rus, Experiments with cooperative networked control of underwater robots, in *Experimental Robotics,* pp. 463–470, 2008.

[34] Hylke W. van Dijk, Collaborative embedded networks for submarine surveillance, Available at: http://cordis.europa.eu/fp7/ict/necs/docs/events/20100602/20100602-02-hwvd-clam_en.pdf [Accessed: 26 October 2010].

[35] N. Abaid and M. Porfiri, *Synchronous dynamics over numerosity-constrained stochastic networks.* Applications of Chaos and Nonlinear Dynamics in Science and Engineering - Vol. 2 Understanding Complex Systems, pp 95-121, 2012.

[36] I. Belykh, M. Di Bernardo, J. Kurths and M. Porfiri, Evolving dynamical networks, *Phys. Nonlinear Phenom.,* vol. 267, pp. 1–6, 2014.

[37] P. DeLellis, M. DiBernardo and M. Porfiri, Achieving consensus and synchronization by adapting the network topology, 18th IFAC World Congress; Milano; Italy; 28 August 2011, 2 September 2011, vol. 18, pp. 1243–1248.

[38] E. S. Larrucea, *Cooperative localization in wireless networked.* ProQuest, Available at: http://search.proquest.com/docview/304810383 2007. [Accessed: 8 November 2013].

[39] Z. Liu, M. Kwiatkowska and C. C. Constantinou, A self-organised emergent routing mechanism for mobile ad hoc networks, *Eur. Trans. Telecommun.,* vol. 16, n. 5, pp. 457–470, 2005.

[40] P. Delellis, M. Dibernardo, F. Garofalo and M. Porfiri, Evolution of complex networks via edge snapping, *IEEE Trans. Circuits Syst. Regul. Pap.,* vol. 57, n. 8, pp. 2132–2143, 2010.

[41] J. Partan, J. Kurose and B. N. Levine, A survey of practical issues in underwater networks, *ACM SIGMOBILE Mob. Comput. Commun. Rev.,* vol. 11, n. 4, pp. 23–33, 2007.

[42] J. Pearl, *Probabilistic reasoning in intelligent systems: networks of plausible inference.* Springer Tracts in Advanced Robotics, Vol. 55, Morgan Kaufmann, 1988.

[43] S. Taraglio and A. Zanela, Cellular neural networks: a genetic algorithm for parameters optimization in artificial vision applications, Proceedings of the 1996 4th IEEE International Workshop on Cellular Neural Networks, and Their Applications, CNNA-96; Seville; Spain; 24 June 1996, pp. 315–320.

[44] C. W. Reynolds, Flocks, herds and schools: a distributed behavioral model, *ACM SIGGRAPH Comput. Graph.,* vol. 21, n. 4, pp. 25–34, 1987.

[45] J. Callmer, M. Skoglund and F. Gustafsson, Silent localization of underwater sensors using magnetometers, *EURASIP J. Adv. Signal Process.,* vol. 2010, n. 709318, pp. 1687–1696, 2010.

[46] C. Detweiler, J. Leonard, D. Rus and S. Teller, Passive mobile robot localization within a fixed beacon field, *Algorithmic Found. Robot. VII,* pp. 425–440.

[47] K. Iswandy, S. Carrella and A. König, Intelligent magnetic sensing system for low power WSN localization immersed in liquid-filled industrial containers, *Knowl.-Based Intell. Inf. Eng. Syst.,* pp. 361–370, 2010.

[48] G. C. Karras and K. J. Kyriakopoulos, Localization of an underwater vehicle using an IMU and a laser-based vision system, in *Proc. of the 15th Mediterranean Conference on Control and Applications,* 2007.

[49] C. M. S. J. Leonard and A. A. B. Shaw, Concurrent Mapping and Localization for Autonomous Underwater Vehicles, presented at Proc. Int. Conf. Field and Service Robotics, Available at: http://cml.mit.edu/~jleonard/pubs/udt97.pdf 1997. [Accessed: 8 November 2013].

[50] A. K. Othman, GPS-less localization protocol for underwater acoustic networks, *Int. J. Comput. Sci. Secur. IJCSS,* vol. 1, n. 1, p. 34.

[51] S. Pagnottelli, S. Taraglio, P. Valigi and A. Zanela, Visual and laser sensory data fusion for outdoor robot localisation and navigation, 12th

International Conference on Advanced Robotics, 2005. ICAR '05; Seattle, WA; United States; 18 July 2005 - 20 July 2005;, vol. 2005, pp. 171–177.

[52] C. Stachniss, Mapping and localization in non-static environments, *Robot. Mapp. Explor.*, pp. 161–175, 2009.

[53] Teller, Localization , Available at: http://courses.csail.mit.edu/6.141/spring2010/pub/lectures/Lec08-Localization.pdf 2010. [Accessed: 3 March 2011].

[54] A. Bahr and J. Leonard, Cooperative localization for autonomous underwater vehicles, in *Experimental Robotics*, pp. 387–395, Available at: http://hdl.handle.net/1721.1/55326. [Accessed: 3 March 2011].

[55] J. J. Leonard and A. Bahr, Cooperative localization for autonomous underwater vehicles , 2009. Available at: https://mit.dspace.org/openaccess-disseminate/1721.1/58207. [Accessed: 3 March 2011].

[56] C. Lytridis, G. S. Virk and E. E. Kadar, Co-operative smell-based navigation for mobile robots, *Climbing Walk. Robots*, 2005,pp. 1107–1117.

[57] S. Nawaz, M. Hussain, S. Watson, N. Trigoni and P. N. Green, An underwater robotic network for monitoring nuclear waste storage pools, *1st Int. ICST Conf. Sens. Syst. Softw. SCUBE*, pp. 236–255, 2009.

[58] D. J. Pack, P. DeLima, G. J. Toussaint and G. York, Cooperative control of UAVs for localization of intermittently emitting mobile targets, *IEEE Trans. Syst. Man Cybern. Part B Cybern.*, vol. 39, n. 4, pp. 959–970, 2009.

[59] A. Aboshosha and A. Zell, Disambiguating robot positioning using laser and geomagnetic signatures, in *Proceedings of the 8th Conference on Intelligent Autonomous Systems IAS*, 2004, vol. 8, pp. 10–13.

[60] A Complete Underwater Electric and Magnetic Signature Scenario Using Computational Modeling Available at: www.beasy.com/pubblications/papers/marelec06.pdf. [Accessed: 3 March 2011].

[61] J. Haverinen and A. Kemppainen, A global self-localization technique utilizing local anomalies of the ambient magnetic field, in *Proceedings of the 2009 IEEE International Conference on Robotics and Automation*, 2009, pp. 4459–4464.

[62] High-resolution marine magnetic surveys for searching underwater cultural resources, Annals Of Geophysics, Vol. 49, N. 6, December 2006.

[63] Desert Star Systems, LLC — SeaTag-GEO Archival Fish Tag. Available at: http://www.desertstar.com/Products_product.aspx?intProductID=7. [Accessed: 17 December 2010].

[64] Archival Fish Tag with RF Transmitting Capabilities. Available at: http://www.tempsensornews.com/biomed/archival-fish-tag-with-rf-transmitting-capabilities/. [Accessed: 1 March 2011].

[65] M. Birsan, Electromagnetic source localization in shallow waters using Bayesian matched-field inversion, *Inverse Probl.*, vol. 22, p. 43, 2006.

[66] WFS Technologies. Available at: http://www.wfs-tech.com/index.php/products/seatooth/. [Accessed: 3 March 2011].

[67] API srl. Available at: http://www.api-automation.it/template.php?pag=62606. [Accessed: 3 March 2011].

[68] T. Nara, S. Suzuki and S. Ando, A closed-form formula for magnetic dipole localization by measurement of its magnetic field and spatial gradients, *IEEE Trans. Magn.*, vol. 42, n. 10, pp. 3291–3293, 2006.

[69] E. W. Grafarend and J. Shan, GPS solutions: closed forms, critical and special configurations of P4P, *GPS Solut.*, vol. 5, n. 3, pp. 29–41, 2002.

[70] B. Paláncz, J. L. Awange and E. W. Grafarend, Computer algebra solution of the GPS N-points problem, *GPS Solut.*, vol. 11, n. 4, pp. 295–299, 2007.

[71] L. Liberti, C. Lavor, N. Maculan and A. Mucherino, Euclidean distance geometry and applications, *SIAM Rev.*, vol. 56, n. 1, pp. 3–69, 2014.

[72] J. Kuszewski, M. Nilges and A. T. Brünger, Sampling and efficiency of metric matrix distance geometry: a novel partial metrization algorithm, *J. Biomol. NMR*, vol. 2, n. 1, pp. 33–56, 1992.

[73] R. Davis, C. Ernst and D. Wu, Protein structure determination via an efficient geometric build-up algorithm, *BMC Struct. Biol.*, vol. 10, n. Suppl. 1, p. S7, 2010.

[74] S. Fairchild, R. Pachter, R. Perrin and others, Protein structure analysis and prediction, *Math. J.*, vol. 5, pp. 64–69, 1995.

[75] J. J. Moré and Z. Wu, Distance geometry optimization for protein structures, *J. Glob. Optim.*, vol. 15, n. 3, pp. 219–234, 1999.

[76] A. V. Moere, Time-varying data visualization using information flocking boids, in *Information Visualization, 2004. INFOVIS 2004. IEEE Symposium on*, 2004, pp. 97–104.

[77] J. Cheng, W. Cheng and R. Nagpal, Robust and self-repairing formation control for swarms of mobile agents, in *AAAI*, 2005, vol. 5, pp. 59–64.

[78] W. J. Crowther, Rule-based guidance for flight vehicle flocking, *Proc. Inst. Mech. Eng. Part G J. Aerosp. Eng.*, vol. 218, n. 2, pp. 111–124, 2004.

[79] Determining interaction rules in animal swarms. Available at: http://beheco.oxfordjournals.org/content/21/5/1106.full. [Accessed: 7 November 2014].

[80] R. De Nardi and O. Holland, Swarmav: a swarm of

miniature aerial vehicles, Available at: http://cogprints.org/5569/ 2006. [Accessed: 7 November 2014].

[81] D. Karaboga, An idea based on honey bee swarm for numerical optimization, Technical report-tr06, Erciyes University, Engineering Faculty, Computer Engineering Department, 2005. Available at: http://www-lia.deis.unibo.it/Courses/SistInt/articoli/bee-colony1.pdf. [Accessed: 7 November 2014].

[82] R. C. Eberhart and J. Kennedy, A new optimizer using particle swarm theory, in *Proceedings of the Sixth International Symposium on Micro Machine and Human Science*, 1995, vol. 1, pp. 39–43.

[83] R. Olfati-Saber, Flocking for multi-agent dynamic systems: algorithms and theory, *Autom. Control IEEE Trans. on*, vol. 51, n. 3, p. 401–420, 2006.

[83] R. dell'Erba, Localisation task for underwater swarms with minimum data, in *Proceedings of IARP* Conference "Bio-Inspired Robotics"-Frascati, Italy 14-15 May 2014, pp. 116-124, ISBN 978-88-8286-309-8, Available at http://www.enea.it/it/produzione-scientifica/pdf-volumi/V2014BioInspiredRobotics.pdf [Accessed: 3 November 2014].